\documentclass{article}

\usepackage{PRIMEarxiv}

\usepackage[utf8]{inputenc} % allow utf-8 input
\usepackage[T1]{fontenc}    % use 8-bit T1 fonts
\usepackage{hyperref}       % hyperlinks
\usepackage{url}            % simple URL typesetting
\usepackage{booktabs}       % professional-quality tables
\usepackage{amsfonts}       % blackboard math symbols
\usepackage{nicefrac}       % compact symbols for 1/2, etc.
\usepackage{microtype}      % microtypography
\usepackage{lipsum}
\usepackage{fancyhdr}       % header
\usepackage{graphicx}       % graphics

\usepackage{multirow}
\usepackage{float}
\usepackage{color,soul}

\graphicspath{{media/}}     % organize your images and other figures under media/ folder

\title{Evaluation of Traffic Signals for Daily Traffic Pattern
}

\author{
  Mohammad Shokrolah Shirazi \\
  E. S. Witchger School of Engineering \\
  Marian University \\
  Indianapolis\\
  \texttt{mshokrolahshirazi@marian.edu} \\
   \And
  Hung-Fu Chang \\
  R.B. Annis School of Engineering \\
  University of Indianapolis \\
  Indianapolis\\
  \texttt{hchang@uindy.edu} \\
}

\begin{document}
\maketitle

\begin{abstract}
The turning movement count  (TMC) data is crucial for traffic signal 
 design, intersection geometry planning, traffic flow, and congestion analysis. This work proposes three methods called dynamic, static, and hybrid configuration for TMC-based traffic signals. A vision-based tracking system is developed to estimate the TMC of six intersections in Las Vegas using traffic cameras. The intersection design, route (e.g. vehicle movement directions), and signal configuration files with compatible formats are synthesized and imported into Simulation of Urban MObility (SUMO) for signal evaluation with realistic data. The initial experimental results based on estimated waiting times indicate that the cycle time of 90 and 120 seconds works best for all intersections. In addition, four intersections show better performance for dynamic signal timing configuration, and the other two with lower performance have a lower ratio of total vehicle count to total lanes of the intersection leg. Since daily traffic flow often exhibits a bimodal pattern, we propose a hybrid signal method that switches between dynamic and static methods, adapting to peak and off-peak traffic conditions for improved flow management. So, a built-in traffic generator module creates vehicle routes for 4 hours, including peak hours, and a signal design module produces signal schedule cycles according to static, dynamic, and hybrid methods. Vehicle count distributions are weighted differently for each zone (i.e., West, North, East, South) to generate diverse traffic patterns.  The extended experimental results for 6 intersections with 4 hours of simulation time imply that zone-based traffic pattern distributions affect signal design selection. Although the static method works great for evenly zone-based traffic distribution, the hybrid method works well for highly weighted traffic at intersection pairs of the West-East and North-South zones.
\end{abstract}

% keywords can be removed

\section{Introduction}
\label{sec:intro}
Intersections are focal points of attention for researchers and engineers where multiple roads converge, causing conflict among various road users, such as vehicles and pedestrians. Traffic signals are key in managing road user access, increasing their safety, and improving traffic flow. Although traffic signal scheduling methods can alleviate the problem, they are statically configured based on some traffic data such as vehicle count, direction of flow dominance, and traffic patterns for different times of the day and week ~\cite{Bisht_CIE2022}. Hence, turning movement count (TMC) data is influential for traffic signal design, including phase timing configuration, as it identifies the volume of vehicles moving in different directions at intersections ~\cite{shirazi_mdpi2022}.

There is a comprehensive procedure for designing traffic signals which often starts with data collection (e.g. traffic volume, crash data), analysis of data and intersection such as finding volume for peak hours, and geometric configuration of intersection and its capacity, and finally, ended up with traffic signal timing design within a simulation, testing and optimization~\cite{Qadri_etrr2020}. The data collection step is a fundamental part that can impact other steps based on the quality of information that can be gauged about the corresponding intersection. High-quality data should provide information about the number of vehicles branching in different directions (e.g., going right, left, and straight) from each approaching zone called turning movement count (TMC). The highly accurate TMC data helps to maintain the precise phase timing, including the duration of each traffic light, to make an informed decision regarding traffic management, intersection design, and signal optimization~\cite{shirazi_mdpi2022}.   

While there are different ways to collect TMC data at intersections, they are dissimilar based on accuracy, cost, and scalability. So, they can be grouped into manual, and automated methods based on benefiting the technological solution. Traditionally, people observe each intersection to manually count them which is labor-intensive, and limited for the period that data can be generated. Other traditional methods such as pneumatic road tubes and loop detectors are exposed to high installation and maintenance costs~\cite {balid_its2017}. The more advanced method relies on hardware equipment such as sensors and intelligent software to help with processing the collected raw data and generating meaningful data such as TMC. For example, global positioning system (GPS) data can relay traffic information through vehicle-to-infrastructure (V2I) communication to road units located at intersections and update the signal cycle in real-time~\cite{tian_jte2021}.  

Data analysis helps to determine the order and timing for each traffic flow direction while designing the traffic signals. The most common approach is to plan the fixed order and duration of traffic signal phases and lights for the entire day called the static method. Although the static method benefits from simplicity in design and management, it doesn't incorporate the different traffic situations that might impact the intersection level of service. For instance, construction road zones or crashes on low congested traffic legs can increase the congestion requiring traffic signal reconfiguration. So, the dynamic traffic scheduling method is proposed to better handle the lively real-time traffic situations. While the dynamic scheduling methods better assist with heavy traffic flows, they are not suggested for the low volume traffic as they increase the complexity and communication overhead of traffic signals with road sensor units for real implementation~\cite{shirazi_southeast2024}. As a result, the hybrid traffic scheduling method is proposed in this work to switch between static and dynamic subject to the observed congestion scenario.

In this work, traffic scheduling methods such as static, dynamic, and hybrid are proposed based on turning movement count data. The main motivation for presenting the hybrid approach is supported by the first round of experiments with estimated TMC data through a vision-based tracking system. The trajectories are interpreted as TMC using the longest common subsequence (LCSS) method and they are imported into SUMO to configure traffic signals for dynamic and static methods with realistic traffic scenarios. Since we observe better performance of each method subject to the intensity of traffic volume, we incorporate a traffic generator module within SUMO to investigate the impact of bimodal distribution with different turning movement count patterns representing incoming flow traffic from each zone (e.g., west, north) on traffic signal scheduling methods including dynamic, static, and hybrid methods. Since these methods can complement each other for different traffic scenarios, the hybrid method implies collaboration between static and dynamic methods for the bimodal distribution, and we investigate the effectiveness of the hybrid method for specific TMC patterns.       

More details about the proposed system framework, and experiments can be found in the rest of the article. The remainder of this article is organized as follows: section~\ref{sec:literature} provides insight into related research studies and the contribution of the article, section~\ref{sec:system_overview} presents the proposed system with integral components for the design and evaluation of traffic signals, section~\ref{sec:experiments} provides experimental results, and finally section~\ref{sec:conclusion} concludes the paper.

\section{Literature Survey}
\label{sec:literature}

Due to contemporary progress in artificial intelligence, machine learning, computing hardware devices, and the demand for their usage in smart cities, the intelligent traffic light system has been proposed by some studies~\cite{Desmira_aes2022, rida_2018}. The core part of the system relies on data collection of traffic data (e.g., vehicle arrival, departure) through different sensors and incorporates them into the design of traffic light timings to lessen the queue length and optimize the waiting time of each vehicle. For example, Desmira et al.~\cite{Desmira_aes2022} utilized fuzzy logic within the microcontroller to better accommodate the traffic light concerning the dynamics of the vehicles that meet at intersections. Since multiple sensors are located at each intersection leg, each sensor determines the vehicle queue, and hence the traffic light timing is adjusted after the communication signal is sent to the microcontroller. While most smart scheduling methods allocate privilege to the largest queue to minimize waiting time, some methods consider traffic volume and grant priority to the shortest queue can be beneficial for effective signal design and reducing the waiting time~\cite{rida_2018}. 

Among the advanced methods, computer vision and machine learning play key roles as they process raw data from sensors such as cameras (e.g., drones, traffic cameras), radar, and lidar and perform object identification, such as segmentation and clustering algorithms. The advancement in utilizing artificial intelligence with collecting highly accurate traffic data is suggesting leveraging them for intelligent traffic management in smart cities. The fundamental part of the system relies on data collection of traffic data (e.g., vehicle arrival, departure) through different sensors and incorporates them into the design of traffic light timings to lessen the queue length and optimize the waiting time of each vehicle. Moreover, the recent advances in deep learning methods encourage the usage of computer vision as the accuracy of estimated traffic data has been improved compared to traditional techniques. For instance, vehicle detection and counting using a faster region-based convolutional neural network over 4500 vehicle images has gained more than 91\% accuracy, and the real-time data has been be used to address the traffic congestion through signal automation~\cite{Ubaid_it2022}.

Among the vision-based object detection algorithms, the YOLO (You Only Look Once) algorithm has received high attention due to its real-time capability and high accuracy. The real-time capability can help to optimize vehicular congestion control in traffic signals\mbox{~\cite{Sivaganesan_ics2024}}. Moreover, it can be leveraged to rapidly identify and classify vehicles in live video feeds as the traffic management systems can dynamically adjust signal timings, leading to improved traffic flow and reduced congestion. Besides vehicle detection and counting\mbox{~\cite{Jayapradha_ics2024}}, YOLO can identify traffic signs for the intelligent driving systems efficiently due to its real-time performance\mbox {~\cite{Song_mdpi2023}}. Although deep learning-based vehicle detection techniques have been widely used by researchers and engineers, there is a lack of studies that use the vision information to collect turning movement count data and develop an efficient traffic signal scheduling based on it.

Traditionally, the order and duration of traffic signal lights for each turning direction are predetermined based on collected historical transportation data, and the corresponding phase timing is statically configured for each light duration (e.g., red, green). Since traffic congestion normally occurs during a daily peak time (e.g., 7:00-9:00 am),  other timing plans will be used for morning and evening peak hours, with longer duration of green times for highly congested turning ways. The static methods benefit from simplicity in design and management since their signal timings are homogeneous and expected by drivers~\cite{ahmed_icce2018}.   

An efficient way to schedule traffic signals is to split the traffic time cycle into multiple phases, where each phase implies flow directions with no conflicts. Hence, the amount of green time for a phase is determined by finding the maximum number of counts among non-conflicting ways and considering that as a proportion of to cycle time for configuring the phases. The total cycle time of the intersection can be configured based on different factors such as total volume, intersection geometry, pedestrian and cyclist timing, and clearance time~\cite{zhuang_aap2018, shirazi_mdpi2022}.

Although the static signal scheduling methods are simple and cost-efficient, they fail to adapt to real-time traffic conditions, and their performance gets worse for highly deviated traffic patterns. The dynamic traffic scheduling methods adjust the timings adaptively based on the current traffic data that they receive from different sensors in real-time. Due to the diversity of the traffic scenarios, Tang et al.~\cite{tang_its2020} proposed a route guidance method with adaptive learning ability from traffic. The learning ability is based on analyzing the traffic conditions for different traffic scenarios and assessing the road congestion index from a system perspective.  Due to the efficiency of reinforcement learning (RL) methods in learning and adapting to complex, dynamic environments, they are often used in some studies for traffic signal control. For instance,~\cite{joo_elsevier2020} utilized Q-learning based on considering the standard deviation of queue lengths, and they demonstrated good performance in terms of queue length and waiting period. Moreover, Sun et al.~\cite{Sun_dynamic2024} proposed a reward function and a cost model to ensure a fair scheduling plan to simplify the decision-making process and improve intersection throughput. Since neglecting fairness scheduling may cause long waiting times for some vehicles,  a traffic signal control is proposed with a fairness scheduling capability dealing with diversity-limited traffic data~\cite{Du_ieee24}.

While dynamic traffic signal controller techniques have gained more popularity recently, they are costly and lack practical implementation
for real traffic scenarios. Recent research studies imply that static signal controller methods are efficient when the intersection is not crowded, especially when the ratio of turning movement count over the number of lanes of
intersection is low. When the intersection gets crowded, the dynamic method is superior and it manages the heavier flow better by allocating effective green
time and reducing the overall normalized waiting times~\cite{shirazi_southeast2024}. So, we propose a hybrid method to effectively utilize dynamic and static methods for on-peak and off-peak traffic for a typical day.  

A typical daily flow distribution mimics bimodal distribution since it has two main peaks representing congested traffic flow in the morning and late afternoon. In this work, we investigate the bimodal distribution with different turning movement count patterns representing incoming flow traffic from each zone (e.g., west, north) for traffic signal scheduling methods including dynamic and static methods. Since these methods can complement each other for different traffic scenarios, we propose a hybrid method that provides collaboration between static and dynamic methods for the bimodal distribution, and we investigate the effectiveness of the hybrid method for specific TMC patterns.       

The contribution of this paper is multi-folded:

\begin{enumerate}
\item The system framework is proposed to incorporate turning movement count data estimated from traffic cameras into SUMO.
\item The traffic scheduling methods based on TMC data are proposed, called static and dynamic methods, and a reinforcement-based method based on TMC data is implemented for comparison purposes. Based on the preliminary experimental results, the hybrid method is proposed to trade off simplicity and busy level of the intersection for traffic flow improvement.
\item In order to evaluate the proposed hybrid method with long-term daily movement count, the proposed system framework generates TMC data with a bimodal distribution to evaluate the effectiveness of the traffic signal scheduling methods.
\end{enumerate}

\section{System Framework}
\label{sec:system_overview}
\begin {figure}[t]
\centering
\includegraphics[scale=0.35]{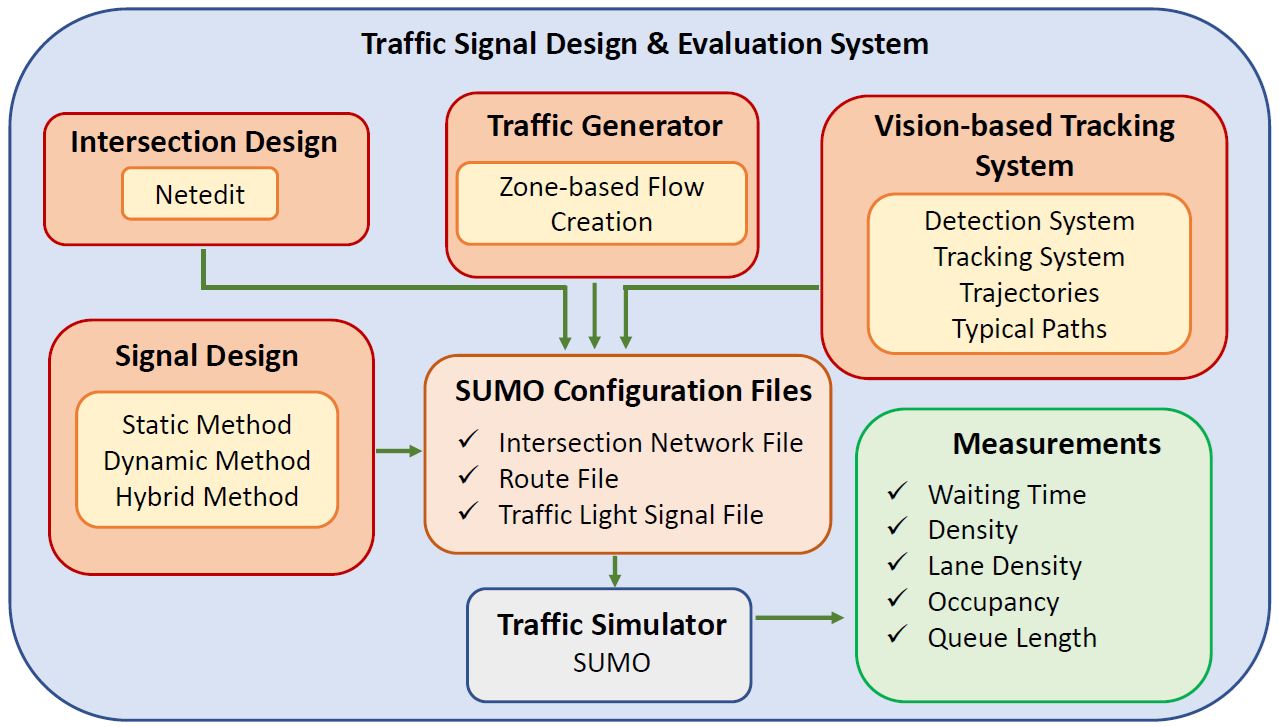}
\caption{The proposed framework for signal design evaluation with zone-based traffic flow generator. }\label{fig:proposed_framework}
\end {figure}

The proposed system framework for the traffic signal evaluation system consists of four core modules namely intersection design, traffic generator, vision-based tracking system, and signal design (See Figure~\ref{fig:proposed_framework}).  These modules generate sumo configuration files including network files, route files, and traffic light signal files as input to the traffic simulator. Finally, the traffic simulator generates the measurements to evaluate the traffic signal and analyze the intersection.

\subsection{Intersection Design}
\label{subsec:intersection_design}
The intersection network file including nodes and roads (i.e., links) is designed using the NETEDIT tool which is a graphical software that allows users to create, and modify road networks in a friendly manner. The intersection network file can be saved in a SUMO-compatible file for later usage in simulations.

\subsection{Traffic Generator}
\label{subsec:traffic_generator}
To mimic the daily traffic flow pattern, the traffic generator module generates turning movement count data for on-peak and off-peak hours according to count average and standard deviations given as inputs to the module. Moreover, it is essential to provide an arbitrary distribution $[w_{1},w_{2},w_{3},w_{4}]$ to the module where $w_{1}+w_{2}+w_{3}+w_{4}=1$ and each $w_{i}$ represents distribution weight for incoming vehicles flowing from each zone of West, North, East, and South. For example, if 100 vehicles are generated for 5:00-6:00 pm, and distribution is defined as $[0.25,0.25,0.25,0.25]$ 25 vehicles from each zone of west, north, east, and south will injected into the road network for same time slots, during that one-hour simulation. The output of the traffic generator is SUMO compatible file presenting the traveling path according to vehicles turning movements simulation time that need to be injected into the network. Moreover, turning movement count data is generated in aggregate for every minute that can be later used for traffic signal light design.       

\subsection{Vision-based Tracking System}
\label{subsec:vision-based tracking system}
\begin {figure}[t]
\centering
\includegraphics[scale=0.35]{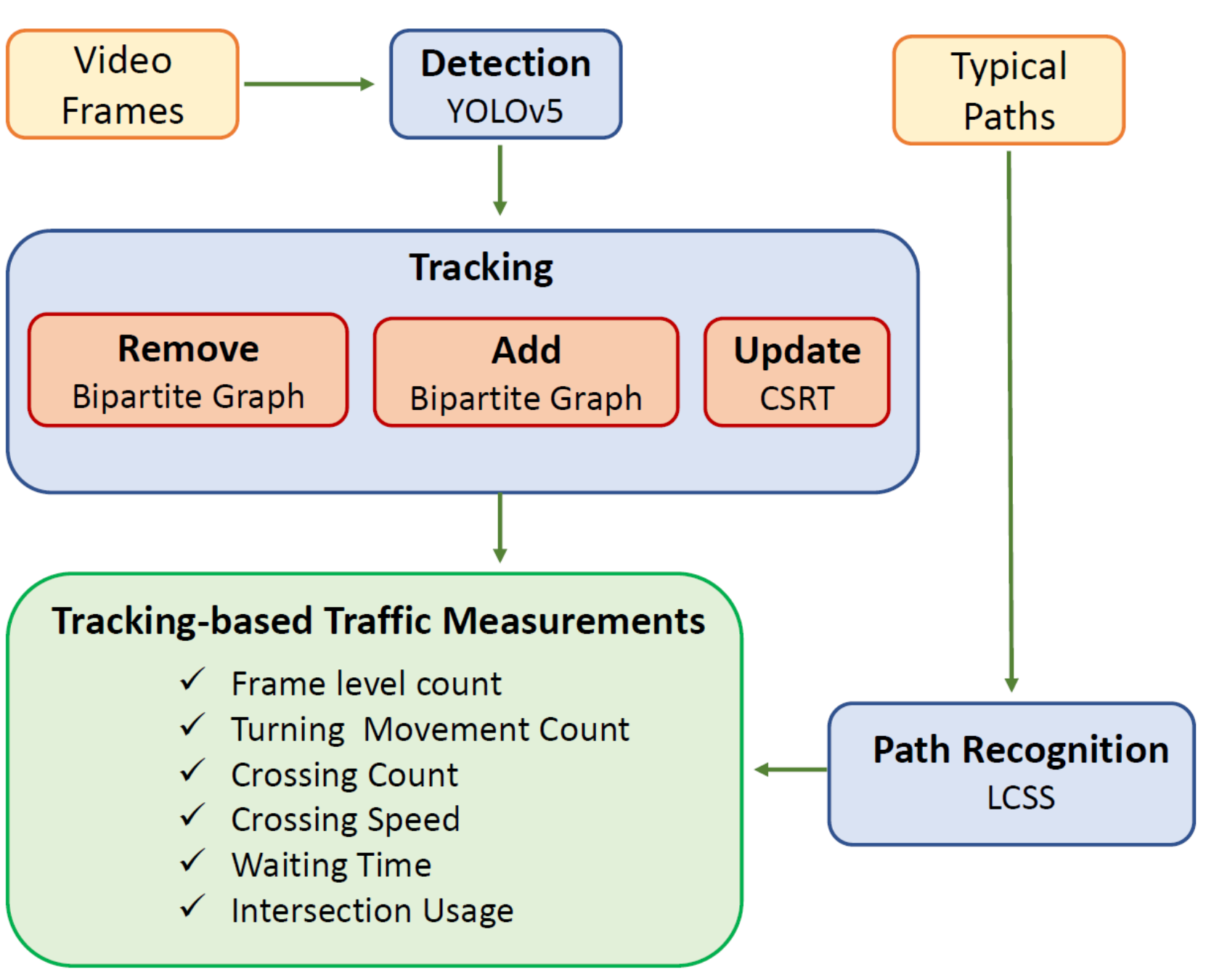}
\caption{The vision-based system overview for intersection analysis.}\label{fig:vision_system_overview}
\end {figure}

To conduct the preliminary intersection analysis, a vision-based tracking system is developed to collect the vehicle trajectories. The proposed system consists of detection and tracking components, and trajectory analysis components are further applied for intersection analysis purposes. Figure\mbox{~\ref{fig:vision_system_overview}} depicts the vision-based tracking system components, which are capable of tracking a mix of road users (e.g., vehicles, pedestrians) and estimating various traffic measurements. However, the vehicle tracking and turning movement count data estimation have been utilized in this work.

\subsubsection{Detection System}
The detection system is the core part of the vision-based tracking system since it feeds the entire system by segmenting road users from the video frames. The convolutional neural network (CNN) features are utilized to predict the bounding box surrounding the object and classify it as a vehicle. The deep neural network uses You Only Look Once (YOLO) Version 5~\cite{YOLOv5} architecture to leverage various informative CNN-based features (e.g., edges, corners, textures) with high speed without sacrificing accuracy.         

\subsubsection{Tracking System}
The tracking system receives the classified road users from the detection system as list of road users with bounding box coordinates for each video frame called  \texttt{DL}. By processing the  \texttt{DL} of each sequence of two frames, it finds the matching road users based on bounding box coordianets as well as appearance similarity using discriminative correlation filter tracking (i.e., \texttt{CSRT()}~\cite{Lukezic_CSRT2016}). The main functionality of the tracking system is object association and create a list of trajectories as $(x,y)$ sequence for each road user.  So, the main process includes updating the bounding box coordinates and using a bipartite graph for association to find a correlation between the updated tracking list of road users and detection items of new road users from incoming frame.

Intersection over union (\ref{eq:IoU}) is an important criterion used by the tracking system to find match. If the obtained numbers is greater than fifty percent (i.e., $IoU>0.50\%$), it will be considered as match. So, the mean of the coordinates are plugged in to the trajectory list ($TL$)if this criterion is met between bounding boxes coordinates of a track and detection. Otherwise, only updated information as output of CSRT() filter will be added for nnex few frames if the detection match is not found before triggering the track removal. This assists with keeping track of temporary losing road users due to noisy situations such as light, illumination change, and occlusions.

\begin {equation}
\label{eq:IoU}
IoU(TL,DL)=\frac{TL\cap DL }{TL \cup DL }
\end {equation}
 
Another case might happen is when the road user bounding box in the \texttt{DL} does not find the existing tracks in the track list during the association step of the algorithm. It is logical to interpret this as the entrance of the new road user to the camera field of view and the algorihm will apply its new bounding box coordinates to the \texttt{TL}.

\subsubsection{Trajectories}
The trajectories are sequence of $(x,y)$ coordinates attach to each road user in $TL$ that is saved by the tracking system at the end of the tracking into a file. Each road user has been labeled by a user id (e.g., 0: pedestrian, 1: vehicle)

\begin{figure}[t]
\centering
\includegraphics[scale=0.25]{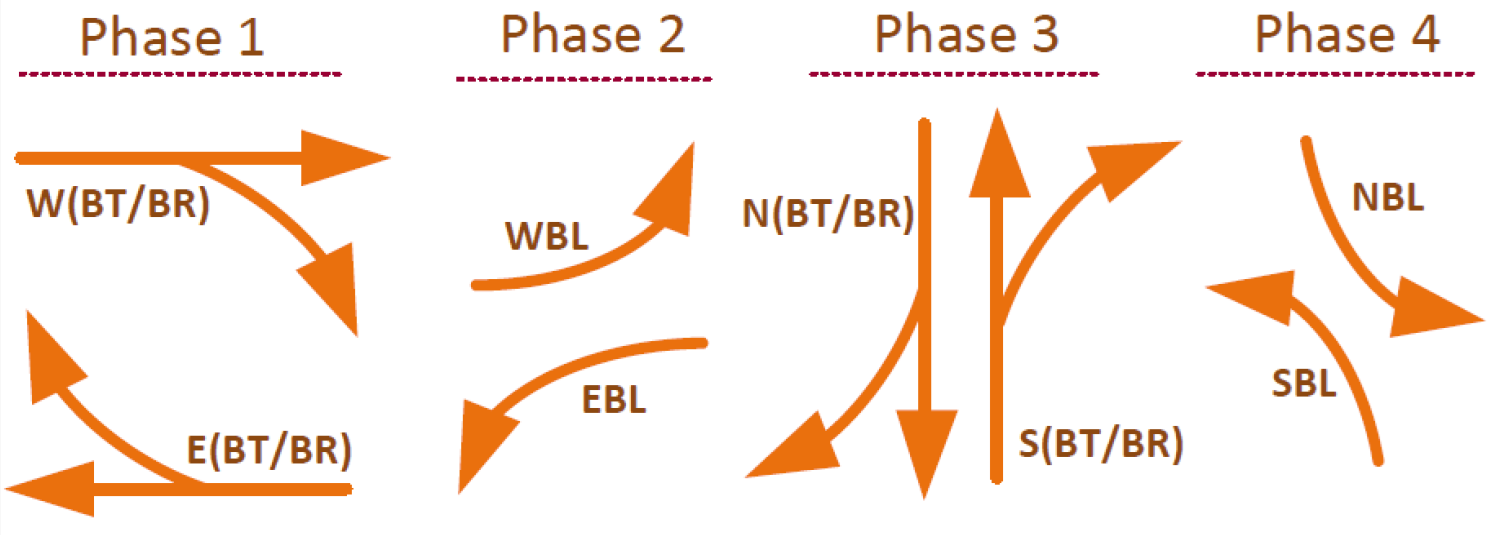}
\caption{Four major signal phases. Left turns of phases 2 and 4 are protected and there are permissive left turns associated with phases 1 and 3.}\label{fig:Signal_design_default}
\end{figure}

\subsubsection{Typical Paths}
The typical paths are utilized by the tracking system to identify turning movement count from the collected trajectories. So, a typical intersection should have 12 typical paths (e.g., WBL) presenting trajectories over the image frame for the intersection. The tracking system pos processes the trajectories and utilizes LCSS method to find the turning movement count. LCSS method is used in the system due to its resilience to disturbances, outliers, and reliable performance for two unequal paths \cite{shirazi_mvap2019}.

\subsection{Signal Design}
\label{subsec:traffic signal design}
After the creation of a road with junctions representing an intersection, the traffic signal can be applied with a configuration file showing the traffic control schedule with timings for each road direction as a sequence of phases. 
The contents of the configuration file are based on TMC data that have been obtained from the traffic generator module. Three different methods are used for the static, dynamic, and hybrid methods of traffic signal design.

\subsubsection{Static Method}
The static method has preconfigured time slots for four major signal phases (Please see the Figure \mbox{\ref{fig:Signal_design_default}}) for the entire simulation time. The green signal times are calculated based on the maximum flow of each concurrent turning movement direction of each phase. For example, in one cycle of 90 seconds, there will be 4 phases of green phases (i.e., $20 \times 4=80 s$) interleaving with 4 phases of yellow lights (i.e., $3 \times 4=12 s$). The same cycle is repeated for the entire simulation time.

\subsubsection{Dynamic Method}
The dynamic method allocates green signal time based on TMC data obtained from different directions according to each traffic signal phase. So, the dedicated green and red light duration is subject to the critical count since the algorithm assesses the TMC data of each non-conflicting direction and compares them to select the largest one as critical count data for green time slot assignment. Figure \ref{fig:Signal_design_default} depicts the four major phases that are fundamental for the design of traffic lights at intersections. Each phase has two major non-conflicting movement directions, and the algorithm selects the critical one based on TMC data for the predefined interval/phase period (e.g., 90 seconds).

The main goal of traffic signals is to supply a fair amount of devoted time to each flow direction based on their requirement. For example, [\texttt{max((WBT+WBR)/2,(EBT+EBR)/2), max(WBL,EBL), max((NBT+NBR)/2, (SBT+SBR)/2), max(NBL, SBL)}] represent four critical TMC as a list of [a,b,c,d]. For a cycle time (e.g., 90 seconds) and total = a+b+c+d, the new green time duration for phase1 would be as follows: \texttt{phase1= (a / total) * cycle time - yellow time}. The yellow time is set with a constant value for all phases (e.g., 3 seconds). Similarly, three other green time assignments are calculated for b, c, and d components of the critical data count. So, eventually, 8 traffic phases will be configured where four of them are yellow phase times after each computed phase green time.

\subsubsection{Reinforcement Learning}
 Reinforcement Learning (RL) offers a transformative solution by enabling adaptive, real-time control that optimizes traffic flow based on live conditions. RL agents learn from experience, continuously improving decisions to minimize delays, prioritize high-traffic directions, and respond to emergencies or unexpected events. Unlike rule-based systems, RL handles complex, nonlinear traffic dynamics, making it ideal for modern smart cities.

  The RL-based traffic light control system leverages Deep Q-Networks (DQN) to allocate green light durations across four directions (WB, NB, EB, SB) based on the provided TMC data. The core innovations contain :
\begin{itemize}
\item Data-Driven States: Traffic volumes are normalized and aggregated from lane-specific TMC data (e.g., WB = WBL + WBT + WBR), providing a compact state representation.
\item Delay Minimization: The reward function penalizes total delay, encouraging the agent to reduce congestion. Delays are calculated as traffic volume divided by allocated green time.
\item Exploration-Exploitation Balance: The agent starts with high randomness (epsilon = 1.0) and gradually shifts to learned strategies, ensuring robust performance in diverse scenarios.
\end{itemize}

In our architecture, the raw TMC data (per minute counts for 12 lane movements) represent an environment that aggregates lane movements into 4 directions and normalizes volumes to [0, 1]. The step function acts as a negative delay and updates the traffic state. The Neural Network is a 3-layer DQN with ReLU activation and softmax output (ensures allocations sum to 1) with input of 4D state (normalized WB, NB, EB, SB volumes), and output of green time allocation percentages.

\subsubsection{Hybrid Method}
The hybrid method takes advantage of static methods and dynamic methods for the off-peak and on-peak hours for a typical bimodal traffic distribution. According to the study in ~\cite{shirazi_southeast2024} for lower congested intersections, the static method is employed, and when congested, the dynamic method becomes effective to handle traffic based on turning movement data. In summary, the predefined time scheduling is generated for the low-congested time of the simulation time, and varied timings for each phase of traffic signals based on TMC are generated for the congested time according to static and dynamic scheduling methods, respectively.

\subsection{SUMO Configuration Files}
\label{subsec:configuration_files}
The SUMO configuration file contains all inputs required by the traffic simulator that is generated by the core modules (i.e., intersection design, traffic generator, signal design) such as the intersection network file, route file showing the traveling path for each turning movement with departing time, and traffic signal configuration files for static, dynamic, and hybrid methods. These files are generated in the appropriate format and compatible with the traffic simulator.

\subsection{Traffic Simulator}
\label{subsec:traffic_simulator}
The traffic simulator is employed to evaluate the traffic signal scheduling methods for a given traffic flow and assess the intersection utilization. The traffic simulator requires various input files as configuration files to run the simulation process by injecting vehicles into the intersection network. By the end of the simulation process,  the output measurements are generated based on the identified measurements provided for the traffic simulator at the beginning of the simulation process.

\subsection{Measurements}
\label{subsec:measurements}
The measurements assist in understanding the behavior of the intersection and the performance of traffic signals for the given scheduling strategies. The waiting time and queue length are important measurements to identify the bottlenecks at intersections and evaluate traffic signal scheduling methods. As a result of this, the important measurements are generated after the simulation is complete.

\section{Experimental Results}
\label{sec:experiments}
The Python programming language is utilized for the implementation of the proposed system including traffic generator and signal design methods due to its open-source feature and extensive libraries and packages. The generated output files such as TMC estimates of a day with bimodal pattern for sorted routes as well as dynamic, static and hybrid traffic light timings are in extensible markup language (XML) formats that can be readily incorporated into Eclipse SUMO 1.13.0 for the simulation and estimating traffic measurements. The system was run on a PC with a quad core Intel i7-7700 3.6 GHz processor, 16 GB RAM and an Nvidia Geforce GTX 1070 GPU with 8 GB of total RAM and 2048 CUDA cores.

\subsection{Intersection Design}
The six intersections of the city of the Las Vegas  were chose since they have their traffic cameras available that allow us to design the intersection networks within the NETEDIT software properly. Thanks to the Freeway and Arterial System of Transportation (FAST) for making the cameras public available, the intersections INT3405 (INT1), INT3428 (INT2), INT3479 (INT3), INT3148 (INT4), INT3243 (INT5), INT3257 (INT6) were selected for our study at the locations described in Table \ref{tab:int_cam_location}.

\begin{table}[H]
\centering
\caption{Intersection locations for traffic monitoring and signal evaluation.}
\begin{tabular}{p{2 cm}p{2cm}p{8 cm}}
    \toprule
     \textbf{INT(Cam)} & \textbf{Time} & \textbf{Location} \\
    \midrule
   INT1(3405) & 4:00-5:00 pm & W Sahara Avenue  at S Durango Drive\\
   INT2(3428) & 4:00-5:00 pm &  W Sahara Avenue at S Cimmaron Drive \\
   INT3(3479) & 4:00-5:00 pm &  W Sahara Avenue at Buffalo W \\
   INT4(3148) & 7:00-8:00 am &   S Grand Central Parkway at Bonneville \\
   INT5(3243) & 7:00-8:00 am & W Sahara Avenue at S Valley View Boulevard \\
   INT6(3257) & 7:00-8:00 am & W Charleston Boulevard at S Grand Central Parkway \\
   \bottomrule
  \end{tabular}\label{tab:int_cam_location}
\end{table}

\begin{figure}[H]
\centering
\includegraphics[scale=0.4]{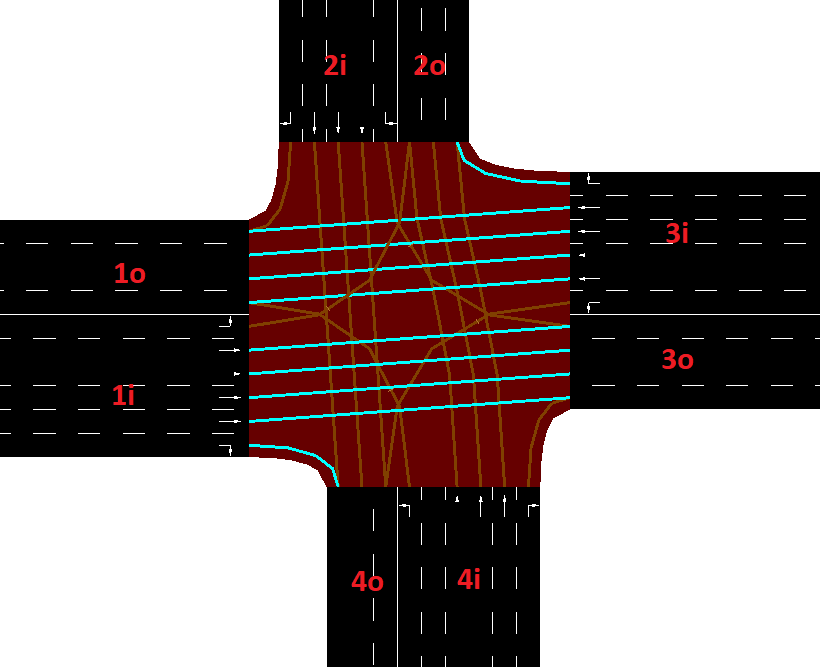}
\caption{Label assignments for network edges (INT1). }\label{fig:INT1_lables}
\end{figure}

 Using NETEDIT software, the intersections are designed by incorporating the number of lanes for each zone direction based on our examination of traffic cameras, and google map. Figure \ref{fig:INT1_lables} presents new labels assigned to each intersection edge for INT1. The  West (W), north (N), east (E), and south (S) have been defined as zone regions using labels 1, 2, 3, and 4. The inflow and outflow directions have been identified using i and o for each edge of directions.

Table \ref{tab:int_design} shows the number of created lanes for each edge of the network for six selected intersections for signal evaluation. For inflow directions (i.e., 1i, 2i, 3i, 4i) to the junction, INT1 has the highest number of lanes which is 6 for West, 5 for North, 6 for East and South departing zones. Obviously, INT4 has the lowest level of service due to the lower number of discharging lanes with outflow directions (i.e., 1o, 2o, 3o, 4o).

\begin{table}[H]
\centering
\caption{Number of lanes for each edge of intersection network.}
\begin{tabular}{p{0.8cm}p{0.35cm}p{0.35cm}p{0.35cm}p{0.35cm}p{0.35cm}p{0.35cm}p{0.35cm}p{0.35cm}p{0.4cm}}
    \toprule
     \multirow{2}{*} {\textbf{INT}}  & \multicolumn{2}{c}{\textbf{West}} & \multicolumn{2}{c}{\textbf{North}}  &   \multicolumn{2}{c}{\textbf{East}} & \multicolumn{2}{c}{\textbf{South}}  & \multirow{2}{*} {\textbf{Total}} \\
      & \textbf{1i} & \textbf{1o} & \textbf{2i} & \textbf{2o} & \textbf{3i}  & \textbf{3o} & \textbf{4i} & \textbf{4o} & \\
    \midrule
   INT1  & 6 & 4 & 5  & 3 & 6  & 4 & 6 & 3 & 37\\
   INT2  & 5 & 4 & 3 & 2 & 5 & 4 & 3 & 2 & 28\\
   INT3  & 6 & 4 & 5 & 3 & 6 & 4 & 5 & 3 & 36\\
   INT4  & 5 & 2 & 4  & 2 & 4  & 2 & 4 & 2 & 25\\
   INT5  & 5 & 3 & 4 & 3 & 5 & 3 & 5 & 2  & 30\\
   INT6 & 6 & 5 & 5 & 3 & 6 & 3 & 5 & 2 & 35\\
   \bottomrule
  \end{tabular}\label{tab:int_design}
\end{table}

\begin{figure}[H]
\centering
\includegraphics[scale=0.85]{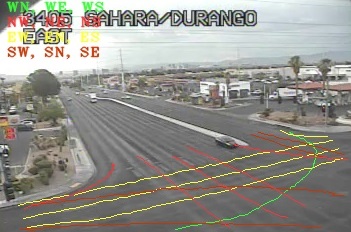}
\caption{Typical paths for INT1. }\label{fig:typical_paths_INT1}
\end{figure}

The first complete trajectories of each moving direction is collected to form a list of typical paths for each intersection as they are used as a reference for generating TMC. Each trajectory can hence be compared with this references to contribute to TMC after tracking of the vehicle. The typical paths for INT1 are shown in Figure~\ref{fig:typical_paths_INT1}. The paths are colored based on zone areas such as green for west zone (e.g., west-to-east (WE) path).

\subsection{Preliminary Results}
\label{sec:pre_results}
The first set of experiments are performed using estimated TMC data through traffic cameras for one hour of observation using the vision-based tracking system. After importing the TMC data as well as intersection designs into SUMO, static and dynamic scheduling methods are evaluated and compared for 6 different intersections with realistic data. The conclusion of the preliminary experiments proposes the idea of hybrid traffic signal methods as it will be explained in Section~\ref{sec:comp_results}.

\subsubsection{TMC Results}
After generating the trajectories using the vision-based tracking system, the TMC is generated shown in Table~\ref{tab:tmc_int}. For the target intersections, all possible movements have been captured except the west branch right and south branch right due to camera limitation in capturing the whole intersection within its field of view.
The blue and red colors are used to highlight the first and second highest counts respectively. The highest number of vehicle movements belong to the west branch through and the left branch, and INT5, INT6 are the busiest.

\begin{table}[H]
\centering
\caption{The turning movement count (TMC) results.}
\begin{tabular}{p{0.6cm}p{0.5cm}p{0.5cm}p{0.5cm}p{0.5cm}p{0.5cm}p{0.5cm}p{0.5cm}p{0.5cm}p{0.5cm}p{0.5cm}p{0.5cm}}
    \toprule
     \multirow{2}{*} {\textbf{INT}}  & \multicolumn{2}{c}{\textbf{West}} & \multicolumn{3}{c}{\textbf{North}}  &   \multicolumn{3}{c}{\textbf{East}} & \multicolumn{2}{c}{\textbf{South}} &\multirow{2}{*} {\textbf{SUM}}   \\
      & \textbf{BL} & \textbf{BT} & \textbf{BL} & \textbf{BT}  & \textbf{BR} & \textbf{BL} & \textbf{BT}  & \textbf{BR}& \textbf{BL} & \textbf{BT} &  \\
    \midrule
   INT1  & 505 & 0   & 233 & \textcolor{blue}{757}  & 214 & 0 & \textcolor{red}{1345} & 10 & 99 &645& 3808\\
   INT2  & 49 & \textcolor{blue}{249}  & 36 & 25 & 39 & 8 & \textcolor{red}{744} & 0 & 40 & 46 & 1236\\
   INT3  & 441 & 1477  & 94 & 1082 & 0 & 157 & 929 & 0 & \textcolor{red}{3186} & \textcolor{blue}{2320} & 6818\\
   INT4  & 133 & 232   & 63 & 277  & 219 & 44 & \textcolor{red}{575} & 0 & 212 & \textcolor{blue}{352} & 2107\\
   INT5  & 0 & 1940  & \textcolor{blue}{2412} & 1249 & 155 & 754 & \textcolor{red}{5049} & 73 & 635 & 2209 & \textcolor{red}{14476}\\
   INT6 & 452 & \textcolor{red}{2232}  & \textcolor{blue}{3153} & 266 & 0 & 497 & 2141 & 558 & 632 & 358 & \textcolor{blue}{10289}\\
   \bottomrule
  \end{tabular}\label{tab:tmc_int}
\end{table} 

The zone capacity rate is defined as the ratio of total vehicle counts on the network edge (e.g. i1, 1o). So, we can calculate it by estimating departing or arriving TMC counts divided by the number of lanes of the network's edge. Table \ref{tab:capacity_rate} presents the zone capacity rates. For instance, the \texttt{C2i=NBL+NBT+NBR} and \texttt{C2o=WBL+EBR+SBT} for the North zone which are 1204 and 1160 for INT1. After dividing them by 5 and 3, 241, and 387 are obtained as shown in Table \ref{tab:capacity_rate} for INT1. This criterion will represent each intersection leg's capacity to handle the traffic requests.

\begin{table}[H]
\centering
\caption{The zone capacity rates.}
\begin{tabular}{p{0.6cm}p{0.45cm}p{0.45cm}p{0.45cm}p{0.45cm}p{0.45cm}p{0.45cm}p{0.45cm}p{0.45cm}p{0.45cm}}
    \toprule
     \multirow{2}{*} {\textbf{INT}}  & \multicolumn{2}{c}{\textbf{West}} & \multicolumn{2}{c}{\textbf{North}}  &   \multicolumn{2}{c}{\textbf{East}} & \multicolumn{2}{c}{\textbf{South}}  & \multirow{2}{*} {\textbf{TC}} \\
      & \textbf{C1i} & \textbf{C1o} & \textbf{C2i} & \textbf{C2o} & \textbf{C3i}  & \textbf{C3o} & \textbf{C4i} & \textbf{C4o} & \\
    \midrule
   INT1  & 84 & 414 & 241  & 387 & 226  & 58 & 124 & 252 & 103 \\
   INT2  & 60 & 206 & 33 & 47 & 150 & 71 & 29 & 16 & 44\\
   INT3  & 320 & 312 & 235 & 920 & 181 & 393 & 528 & 413 & 189\\
   INT4  & 73 & 503 & 140  & 242 & 155  & 147 & 141 & 160 & 84\\
   INT5  & 388 & 1946 & 954 & 761 & 1175 & 1451 & 569 & 1001  & \textcolor{red}{483}\\
   INT6 & 447 & 555 & 684 & 456 & 533 & 1795 & 198 & 381 & \textcolor{blue}{294}\\
   \bottomrule
  \end{tabular}\label{tab:capacity_rate}
\end{table} 

We define the total capacity (TC) rate as the ratio of the total count over the total lanes for an intersection. As you may notice, Branch Thru (BT) and Branch Left (BL) are the highest contributing factors due to the higher number of counts. The INT5 and INT6 are subject to review of traffic signal scheduling and planning the number of lanes. The two lowest capacity rates belong to INT2 and INT4 implying a safe level of service with the current configurations of intersection and traffic signal.    

\subsubsection{Signal Evaluation Results}
The simulation process is conducted after synthesizing the configuration data (i.e., sorted routes, TLS configurations), and network file by SUMO with a time of 1 hour for six intersections. The estimated waiting time is divided by the total number of vehicles required to evaluate signal scheduling methods (e.g., static, dynamic, reinforcement learning) called normalized waiting time~\cite{shirazi_mdpi2022}.

\begin{figure}[H]
\centering
\includegraphics[scale=0.6]{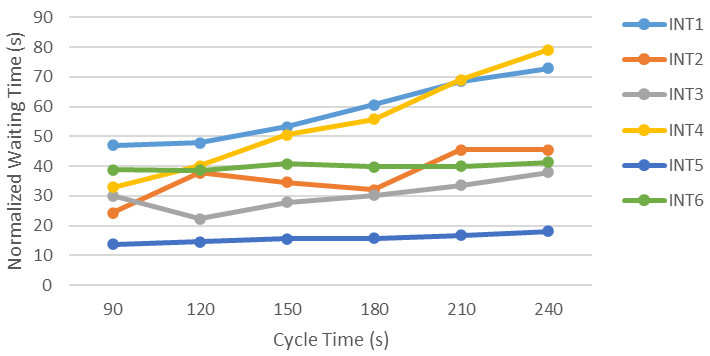}
\caption{The NWT measurements for static scheduling method.  }\label{fig:static_experiments}
\end{figure}

Figure \ref{fig:static_experiments} shows the estimated normalized waiting time (NWT) and suggests that the 90-second cycle has the highest performance for INT1, INT2, INT4, and INT5, while the 120-second cycle has the highest performance for INT3 and INT6. The intersections with low level of service are INT1 and INT6 based on the normalized waiting time, suggesting a redesign of the traffic signal.

\begin{figure}[H]
\centering
\includegraphics[scale=0.6]{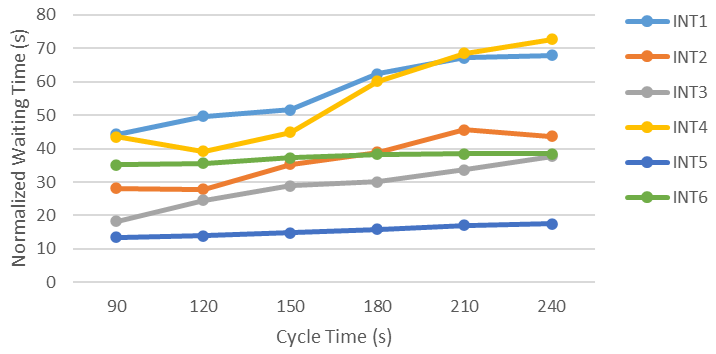}
\caption{The NWT measurements for dynamic scheduling method. }\label{fig:dynamic_experiments}
\end{figure}

The experiments for the dynamic traffic signal scheduling method are presented in Figure \ref{fig:dynamic_experiments}, which advocate better performance for cycle 90 for all intersections except INT2 and INT4 which have a lower NWT. Experiments on both scheduling methods show an increasing trend for a longer cycle time.

\begin{figure}[H]
\centering
\includegraphics[scale=0.95]{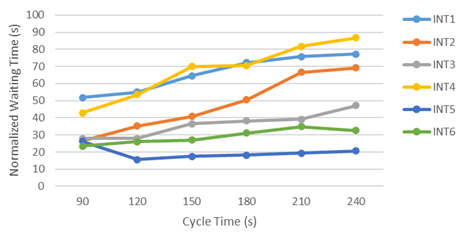}
\caption{The NWT measurements for reinforcement learning scheduling method.}\label{fig:rl_experiments}
\end{figure}

Figure \ref{fig:rl_experiments} depicts the evaluation of the six intersections with the reinforcement learning method while using one hour of TMC data as training samples. Similarly, the RL algorithm shows an increasing trend when the cycle time increases. Moreover, it demonstrates high normalized waiting time for INT1 and the lowest for INT5. However, the algorithm shows superior performance for INT6, which has the second highest TC. This implies that reinforcement learning can also be an alternative to the dynamic method for busy intersections if there is more training data as well as environmental inputs (e.g., queue length) to assist with improving itself over time. 

In order to proceed with a comparison of the proposed signal scheduling methods, the method (i.e. dynamic, static, RL) with the lowest NWT and its corresponding cycle time is selected for each intersection. The comparison results are shown in the Table \ref{tab:signal_methods_compare}.  The dynamic method represents better performance for the INT1, INT3, and INT5; the reinforcement learning method works better for the second busiest intersection (i.e., INT6), while the static method excels for the INT2 and INT4. The results of the zone capacity rate (see Table \ref{tab:capacity_rate}) indicate that INT2 and INT4 have the lowest TC values. 
This supports the fact that the dynamic method is not always great, especially when the ratio of turning movement count to the number of lanes of intersection is low. 

\begin{table}[H]
\centering
\caption{The comparison of signals scheduling methods.}
\begin{tabular}{p{1.2cm}p{0.95cm}p{0.65cm}p{0.65cm}p{0.65cm}p{0.65cm}p{0.65cm}p{0.65cm}}
    \toprule
     \textbf{Method}     & \textbf{Criteria} & \textbf{INT1} & \textbf{INT2} & \textbf{INT3} & \textbf{INT4}  & \textbf{INT5} & \textbf{INT6}  \\
    \midrule
   \multirow{2}{*}  {\textbf{Static}} &  {\textbf{NWT}} & 47.02 & 24.30  & 22.38 & 32.96  & 13.69 & 38.57  \\
     & {\textbf{Cycle}} & 90 & 90 & 120 & 90 & 90 & 120 \\
   \multirow{2}{*} {\textbf{Dynamic}}  & {\textbf{NWT}} & 44.30 & 27.82 & 18.19 & 39.17 & 13.46 & 35.18 \\
     & {\textbf{Cycle}} & 90 & 120  & 90 & 120  & 90 & 90 \\
      \multirow{2}{*} {\textbf{RL}}  & {\textbf{NWT}} & 51.85 & 26.42 & 27.89 & 42.78 & 15.57 & 23.51 \\
     & {\textbf{Cycle}} & 90 & 90  & 120 & 90  & 120 & 90 \\
      \bottomrule
  \end{tabular}\label{tab:signal_methods_compare}
\end{table} 

In order to have a general and comprehensive method to cover the whole day, a mix of static and dynamic methods is suggested. For example, when the intersection is busier, the dynamic method better handles dedicating effective green time to different demands to reduce the normalized waiting times, and for lower congested times, the static method can better handle the traffic.

\subsection{Complementary Results}
\label{sec:comp_results}
The complementary experimental results are carried out to evaluate the proposed hybrid scheduling method for a typical day of traffic. The method switches between static and dynamic methods for the generated count data, which follow a bimodal pattern. 

\subsubsection{Traffic Generator}
\begin{figure}[H]
\centering
\includegraphics[scale=0.4]{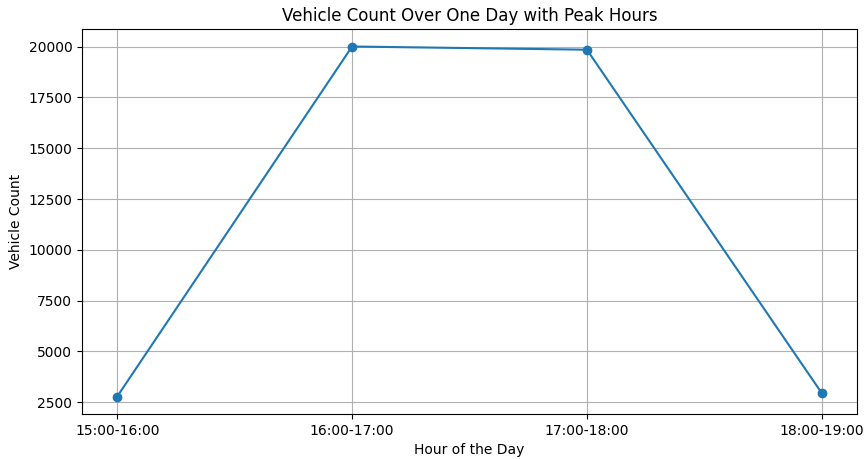}
\caption{The generated vehicle count with a bimodal distribution. }\label{fig:count_bimodal}
\end{figure}

The traffic generator module generates TMC count data for four hours according to the bimodal distribution. Figure \ref{fig:count_bimodal} depicts a typical 4 hour count with a bimodal pattern that is generated and provided to the TMC module. The count data are randomly generated with normal distribution with  $\mu_{a}=2500$,  $\sigma _{a} = 300$ for non-busy time and  $\mu_{b}=20000$, $\sigma _{b} = 400$ for peak time.

\begin{figure}[H]
\centering
\includegraphics[scale=0.72]{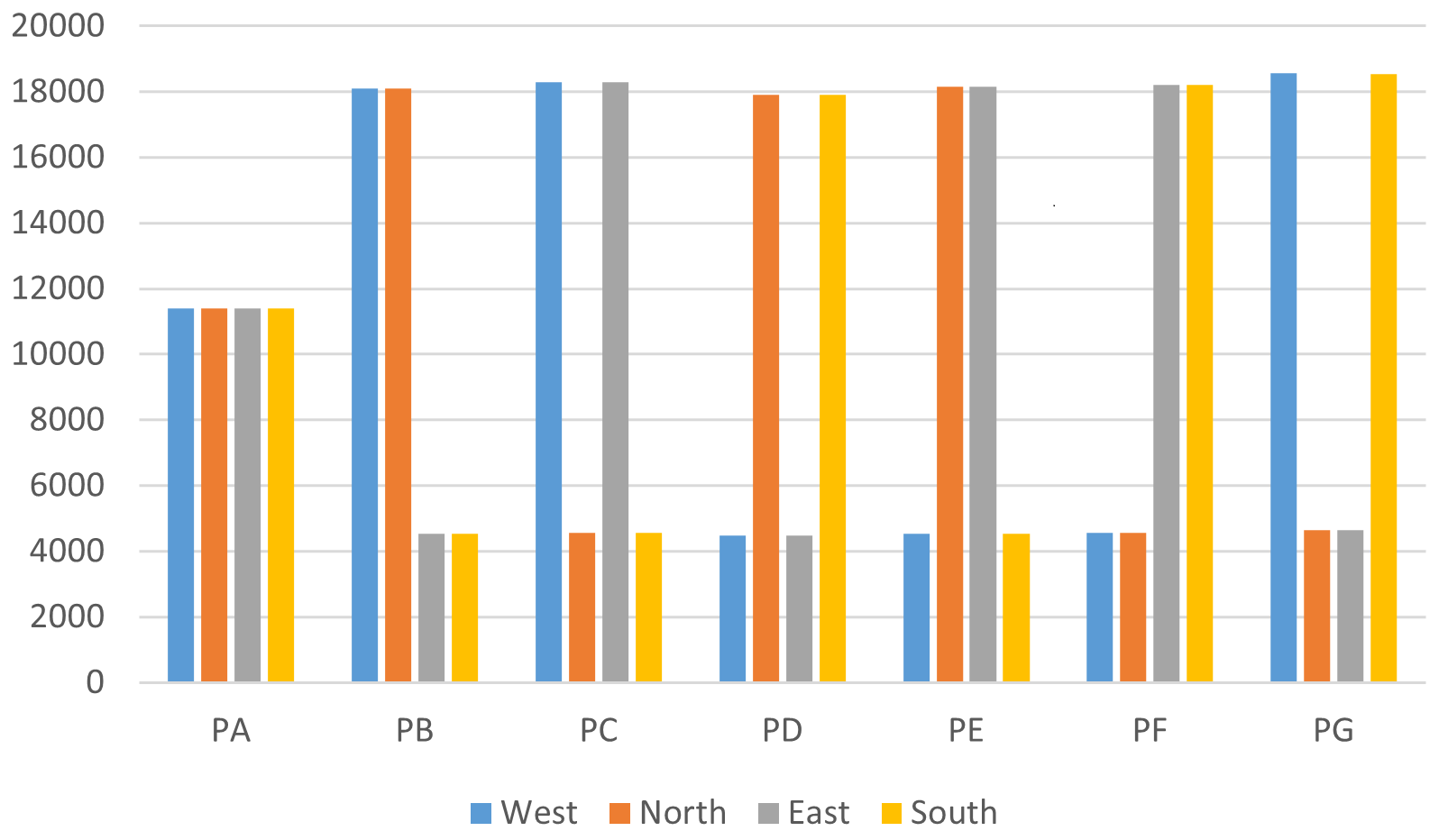}
\caption{The generated count patterns for each zone. }\label{fig:zone_count_pattern}
\end{figure}

The TMC module generates turning movements based on the provided 4-hour total count and weighted patterns for every hour. Finally, the compatible route file is generated for every minute intervals reflecting  4-hour bimodal distribution for each count pattern. Figure \ref{fig:zone_count_pattern} presents the TMC-generated count patterns of four hours by the traffic generator. Seven count patterns show the count weight for each zone (e.g., west, north, east, south). The homogeneous count pattern $PA=\{0.25, 0.25, 0.25, 0.25\}$  indicates an equal number of counts as a portion of the total count for each zone. However, the other base patterns such as $PB$, $PC$, and $PE$ have weights on two different intersection zones. For example, the weight count pattern for $PB$ is $\{0.4, 0.4, 0.1, 0.1\}$, $PC$ is   $\{0.4, 0.1, 0.4, 0.1\}$, and $PE$ is   $\{0.1, 0.4, 0.4, 0.1\}$. Suppose universal count pattern is  $PU = \{0.5, 0.5, 0.5, 0.5\}$ and we define complementary count patterns $PD =  PU- PC$, $PF= PU-PB$, and $PG= PU-PE$.

The main motivation for generating the count patterns is to investigate the effectiveness of the traffic signal methods (e.g. static, dynamic, hybrid) when the count distribution is weighted on different zones even though the total count follows the bimodal distribution.

\subsection{Signal Design \& Configuration Files }
The generated TMC data with bimodal distribution for each count pattern (e.g., $PA$, $PB$) has a departure time instance (i.e. second) and has been labeled with one of the turning movements. The generated route file is formatted in the Extensible Markup Language (XML) format, and assigned with the network edge that can be utilized by the SUMO.

For each traffic signal scheduling algorithm (i.e., static, dynamic, hybrid), the phase timings for every minute time interval are produced which end up to $4 \times 60$ rows for a 4-hour simulation time. For the static algorithm, the contents of each row are identical which ends up with 4 phases or pairs of $(t_{green}, t_{yellow})$ where  $t\_{yellow}$ is 3 and ${t\_green}$ is 20. For the dynamic scheduling algorithm, the ${t\_green}$ is varied for each phase according to the algorithm presented in Section \ref{subsec:traffic signal design}. Finally, the hybrid method utilizes the static configuration for the first and last hour of the simulation and the dynamic one for the busy time during the second and third hours. These traffic signal configurations are prepared and formatted for every cycle time and are imported during the simulation with network and route files.

The simulation by SUMO is run by receiving the configuration data (i.e., turning movement routes, traffic light signal configurations), and network file for four hours for six intersections and three different scheduling methods. The output measurement called waiting time is normalized over the total number of vehicles and it is evaluated for three different methods of static, dynamic and hybrid scheduling methods.  

\subsubsection{Signal Evaluation Results }
Table \ref{tab:signal_design_compare} presents the selected signal scheduling method based on the minimum normalized waiting time (NWT) obtained in experiments. This implies that different traffic pattern models impact the selection of the traffic signal algorithm method. As a matter of fact, all 6 intersections demonstrate different minimum normalized waiting times for different traffic patterns. While INT1 provides more flexibility to apply non-static methods (i.e., hybrid, dynamic) due to its higher total number of lanes, INT2 and INT5 are prone to take the static method as is evident among most traffic pattern models. This supports the fact that dynamic and hybrid methods can all be applicable if their traffic pattern models are not uniform.

\begin{table}[H]
\centering
\caption{Signal design methods evaluation (Static: S, Dynamic: D, Hybrid: H).}
\begin{tabular}{p{0.8cm}p{0.55cm}p{0.55cm}p{0.55cm}p{0.55cm}p{0.55cm}p{0.65cm}}
    \toprule
     \textbf{Pat}     &  \textbf{INT1} & \textbf{INT2} & \textbf{INT3} & \textbf{INT4}  & \textbf{INT5} & \textbf{INT6}  \\
    \midrule
  PA & S & S  & S & S  & S & D  \\
    PB  & D & H & D & S & S & S \\
   PC   & H & S & H & H & H & S \\
    PD  & H & H  & H & H  & H & H \\
    PE  & D & S  & S & S  & S & S \\
    PF  & S & S  & D & H  & S & D \\
    PG  & D & S  & S & S  & S & S \\
      \bottomrule
  \end{tabular}\label{tab:signal_design_compare}
\end{table} 

PA has a uniform distribution pattern and obviously static method is superior. For other patterns and their complementary models, similar evidence is observed for the selection of the traffic signal strategy. $PB$, and $PF$ have at least 3 non-static methods (i.e., $D+H$), $PC$ and $PD$ are the superior models in selecting the hybrid models, and $PE$, and $PG$ are those that are performing well with the static method. This implies the fact that if the count of vehicles is heavier on two opposing inflows (e.g., West-East, North-South), the hybrid methods work well due to the nature of the algorithm in dedicating higher green time to the maximum of these flows.      

\section{Conclusion}
\label{sec:conclusion}
In this paper, a simulation framework is presented to generate realistic TMC data based on daily bimodal distribution and utilize them for different traffic signal evaluation algorithms called static, dynamic, and hybrid methods. The first round of experiments with the TMC data collected through traffic cameras using the proposed vision-based tracking system suggests utilizing the hybrid method to balance the static and dynamic methods for off-peak and peak traffic hours. While the static method always assigns the same green time slots for the phase of the signal, the dynamic method assigns it based on the higher proportion of each turning. The hybrid method utilizes the static method for the off-peak time and the dynamic method for the peak time. Seven different count pattern models are introduced by assigning different weights of vehicle count on each leg of the intersection. The experimental results on six intersections using SUMO show that the count pattern model plays a key role in the selection of the traffic signal scheduling algorithm.

%Bibliography
\bibliographystyle{unsrt}  
\bibliography{references}

\begin{thebibliography}{10}

\bibitem{Bisht_CIE2022}
Abhyudai Bisht, Khilan Ravani, Manish Chaturvedi, Naveen Kumar, and Shailesh Tiwari.
\newblock Indigenous design of a traffic light control system responsive to the local traffic dynamics and priority vehicles.
\newblock {\em Computers \& Industrial Engineering}, 171:108503, 2022.

\bibitem{shirazi_mdpi2022}
Mohammad Shokrolah~Shirazi, Hung-Fu Chang, and Shahab Tayeb.
\newblock Turning movement count data integration methods for intersection analysis and traffic signal design.
\newblock {\em Sensors}, 22(19):7111, 2022.

\bibitem{Qadri_etrr2020}
Syed Shah Sultan~Mohiuddin Qadri, Mahmut~Ali G{\"o}k{\c{c}}e, and Erdin{\c{c}} {\"O}ner.
\newblock State-of-art review of traffic signal control methods: challenges and opportunities.
\newblock {\em European transport research review}, 12:1--23, 2020.

\bibitem{balid_its2017}
Walid Balid, Hasan Tafish, and Hazem~H Refai.
\newblock Intelligent vehicle counting and classification sensor for real-time traffic surveillance.
\newblock {\em IEEE Transactions on Intelligent Transportation Systems}, 19(6):1784--1794, 2017.

\bibitem{tian_jte2021}
Xinmei Tian, Deqi Chen, Xuedong Yan, Liwei Wang, Xiaobing Liu, and Tong Liu.
\newblock Estimation method of intersection signal cycle based on empirical data.
\newblock {\em Journal of Transportation Engineering, Part A: Systems}, 147(3):04021001, 2021.

\bibitem{shirazi_southeast2024}
Mohammad~Shokrolah Shirazi, Hung–Fu Chang, and Mohsen Jahandardoost.
\newblock Evaluation of traffic signal scheduling methods based on turning movement count.
\newblock In {\em SoutheastCon 2024}, pages 504--509, 2024.

\bibitem{Desmira_aes2022}
Desmira Desmira, Mustofa Abi~Hamid, Norazhar~Abu Bakar, Muhammad Nurtanto, and Sunardi Sunardi.
\newblock A smart traffic light using a microcontroller based on the fuzzy logic.
\newblock {\em IAES International Journal of Artificial Intelligence}, 11(3):809, 2022.

\bibitem{rida_2018}
Nouha Rida, Mohammed Ouadoud, Aberrahim Hasbi, and Samira Chebli.
\newblock Adaptive traffic light control system using wireless sensors networks.
\newblock In {\em 2018 IEEE 5th International Congress on Information Science and Technology (CiSt)}, pages 552--556. IEEE, 2018.

\bibitem{Ubaid_it2022}
Muhammad~Talha Ubaid, Tanzila Saba, Hafiz~Umer Draz, Amjad Rehman, Muhammad~Usman Ghani~khan, and Hoshang Kolivand.
\newblock Intelligent traffic signal automation based on computer vision techniques using deep learning.
\newblock {\em IT Professional}, 24(1):27--33, 2022.

\bibitem{Sivaganesan_ics2024}
D.~Sivaganesan, S.~Barath Kumar, S.~Krishnadharani, and G.~Rashmitha.
\newblock Optimization of vehicular congestion control in traffic signals using yolo algorithm.
\newblock In {\em 2024 International Conference on Smart Systems for Electrical, Electronics, Communication and Computer Engineering (ICSSEECC)}, pages 343--348, 2024.

\bibitem{Jayapradha_ics2024}
J.~Jayapradha, R.~Sathishkumar, R.~Swathi, S.~Tejaswini, and J.R. Rinjima.
\newblock Computer vision based enhanced traffic management system using yolo.
\newblock In {\em 2024 International Conference on System, Computation, Automation and Networking (ICSCAN)}, pages 1--6, 2024.

\bibitem{Song_mdpi2023}
Weizhen Song and Shahrel~Azmin Suandi.
\newblock Tsr-yolo: A chinese traffic sign recognition algorithm for intelligent vehicles in complex scenes.
\newblock {\em Sensors}, 23(2), 2023.

\bibitem{ahmed_icce2018}
Eltayeb~KE Ahmed, Amr~MA Khalifa, and Ahmed Kheiri.
\newblock Evolutionary computation for static traffic light cycle optimisation.
\newblock In {\em 2018 International Conference on Computer, Control, Electrical, and Electronics Engineering (ICCCEEE)}, pages 1--6. IEEE, 2018.

\bibitem{zhuang_aap2018}
Xiangling Zhuang and Changxu Wu.
\newblock Display of required crossing speed improves pedestrian judgment of crossing possibility at clearance phase.
\newblock {\em Accident Analysis \& Prevention}, 112:15--20, 2018.

\bibitem{tang_its2020}
Chuanhui Tang, Wenbin Hu, Simon Hu, and Marc~EJ Stettler.
\newblock Urban traffic route guidance method with high adaptive learning ability under diverse traffic scenarios.
\newblock {\em IEEE Transactions on Intelligent Transportation Systems}, 22(5):2956--2968, 2020.

\bibitem{joo_elsevier2020}
Hyunjin Joo, Syed~Hassan Ahmed, and Yujin Lim.
\newblock Traffic signal control for smart cities using reinforcement learning.
\newblock {\em Computer Communications}, 154:324--330, 2020.

\bibitem{Sun_dynamic2024}
Guangling Sun, Rui Qi, Yulong Liu, and Feng Xu.
\newblock A dynamic traffic signal scheduling system based on improved greedy algorithm.
\newblock {\em PLoS one}, 19(3):e0298417, 2024.

\bibitem{Du_ieee24}
Xinqi Du, Ziyue Li, Cheng Long, Yongheng Xing, Philip~S. Yu, and Hechang Chen.
\newblock Felight: Fairness-aware traffic signal control via sample-efficient reinforcement learning.
\newblock {\em IEEE Transactions on Knowledge and Data Engineering}, 36(9):4678--4692, 2024.

\bibitem{YOLOv5}
Glenn Jocher, Alex Stoken, Jirka Borovec, NanoCode012, ChristopherSTAN, Liu Changyu, Laughing, Adam Hogan, lorenzomammana, tkianai, yxNONG, AlexWang1900, Laurentiu Diaconu, Marc, wanghaoyang0106, ml5ah, Doug, Hatovix, Jake Poznanski, Lijun Yu, changyu98, Prashant Rai, Russ Ferriday, Trevor Sullivan, Wang Xinyu, YuriRibeiro, Eduard~Reñé Claramunt, hopesala, pritul dave, and yzchen.
\newblock ultralytics/yolov5: v3.0, August 2020.

\bibitem{Lukezic_CSRT2016}
Alan Lukezic, Tom{\'{a}}s Voj{\'{\i}}r, Luka Cehovin, Jiri Matas, and Matej Kristan.
\newblock Discriminative correlation filter with channel and spatial reliability.
\newblock {\em CoRR}, abs/1611.08461, 2016.

\bibitem{shirazi_mvap2019}
Mohammad Shokrolah~Shirazi and Brendan~Tran Morris.
\newblock Trajectory prediction of vehicles turning at intersections using deep neural networks.
\newblock {\em Machine Vision and Applications}, 30(6):1097--1109, Sep 2019.

\end{thebibliography}

\end{document}